# A Universality-Individuality Integration Model for Dialog Act Classification

Pengfei Gao and Yinglong Ma

School of Control and Computer Engineering, North China Electric Power University, Beijing 102206, China
`yinglongma@ncepu.edu.cn`

**Abstract.** Dialog Act (DA) reveals the general intent of the speaker utterance in a conversation. Accurately predicting DAs can greatly facilitate the development of dialog agents. Although researchers have done extensive research on dialog act classification, the feature information of classification has not been fully considered. This paper suggests that word cues, part-of-speech cues and statistical cues can complement each other to improve the basis for recognition. In addition, the different types of the three lead to the diversity of their distribution forms, which hinders the mining of feature information. To solve this problem, we propose a novel model based on universality and individuality strategies, called Universality-Individuality Integration Model (UIIM). UIIM not only deepens the connection between the clues by learning universality, but also utilizes the learning of individuality to capture the characteristics of the clues themselves. Experiments were made over two most popular benchmark data sets SwDA and MRDA for dialogue act classification, and the results show that extracting the universalities and individualities between cues can more fully excavate the hidden information in the utterance, and improve the accuracy of automatic dialogue act recognition.

## 1 Introduction

Dialog Act (DA) is the smallest unit of language communication, and its interaction actually represents the process of the speaker's language communication [1]. DA represents the illocutionary force of the utterance, which reflects the intention of the speaker to some extent. For example, orders, requests, threats all contain the purpose of making the hearer do something; While making a promise, swearing and other DAs show the intention of taking responsibility for the event [2]. Therefore, the realization of automatic recognition of dialogue act is of great significance to the construction of a perfect dialogue system.

Recently, deep learning and representation learning have achieved good performance in processing natural language processing tasks [2]. As a common scene in natural language processing, Dialog Act Classification (DAC) is the application of these methods. In the DAC, the mainstream one is to continuously extract information helpful for Classification to complete the act recognition of utterance in samples by applying deep learning models, such as Convolutional Neural Network (CNN), Recurrent Neural Network (RNN), etc., by learning generated word vectors or sentence vectors [3, 4].

Although the above mentioned feature extraction scheme based on the vector of words or sentences shows unprecedented advantages, we will inevitably encounter some problems. For example, the dimension of word vector obtained after feature extraction is too high, and the weight of each feature word cannot be calculated directly [5]. The vectors not only bring huge computation amount, but also some feature words in these vectors are actually meaningless to classification [6]. Therefore, it is an urgent problem to estimate the contribution of feature words to the classification system. Considering the characteristics of natural language, feature words containing the key information of utterance are often content words, such as nouns, pronouns, adjectives, etc. Some function ones, such as articles, prepositions and conjunctions, are usually part of some utterances, but they make little contribution to classification [7]. In addition, a sentence often contains many words that have completely different meanings according to their parts of speech. For example, "*bear*" as a verb is often endure and assume (" *He can't bear being laughed at.* "); However, if it's a noun, it means the animal bear (" *He had been mauled by a bear.* "). It may reduce the accuracy of DA recognition if above phenomenon is not dealt with. Therefore, if part-of-speech (POS) cues are taken into account in classification, then more useful features can be extracted to bring greater information gain.

In fact, it is not difficult to find that the illocutionary force contained in a sentence can be actually demonstrated by some linguistic means [8, 9]. In English, for example, these are word order, stress, intonation, tone, punctuation, and so on. Using punctuation as the core example, "*Does not it pretty?* ", " *Does not, it pretty!* " The two sentences are composed of the same words, the former as a rhetorical question and the latter as an exclamatory sentence due to the difference in punctuation. If the proposed classification model does not consider the symbolic cues for such problems, it may not be able to complete the act recognition of the above-mentioned utterances. Besides, the utterances of different lengths tend to have different acts. Most complete sentences should contain the subject, predicate, object and other supplementary components. This complex sentence structure also limits the length of the utterance. This is also generally marked as expounding and expressing some of their subordinate acts. Shorter words like "*OK.*", "*Sure.*" and "*Yes.*" are likely responses to requests, inquiries, etc. These linguistic instruments of illocutionary force are obtained by counting lexical and syntactic information, and we incorporate these features as inputs to the model. In this paper, we collectively refer to this clue as statistics feature.

In this paper, we propose to aggregate word vector, part-of-speech and statistics as the new representation features, which are input into the model to complete the training of the DA recognition model. In our opinion, the simple concatenation of the three features seems to be able to collect all the information of the three vectors, but the model may not be able to explore the key information. Here we argue that their heterogeneity brings about difference in the shape of their distribution as features of different types. However, all of them are derived from the same utterance, so there should be certain connection and similarity between them in theory. Considering that they are originally different features, they should have their own characteristics. If two sub-tasks are constructed, then one is used to learn the relationship between features, so as to reduce the divergence between feature vectors caused by the difference in the distribution form of features. Second, we choose these cues as input features, naturally hoping that they can fully express their own characteristics, so as to achieve a complementary effect with the rest. Then, a space is needed to learn the personality of features, in order to explore the characteristics of different types of features. After the processing of the above tasks, we will gain a new vector aggregating feature commonality and personality, and when it is input into the classifier, the model will achieve more outstanding performance than the simple concatenation. Because the key idea of the model is to explore the Universality and Individuality among features, we called our proposed model the Universality-Individuality Integration Model (UIIM) in this paper.

This work draws on many of the latest technologies in natural language processing, such as BERT (Bidirectional Encoder Representations from Transformers), the Self-Attention mechanism, and BiLSTM (Bi-directional Long Short-Term Memory) and so on, which greatly improves the performance of the model [3-5]. SwDA (Switchboard Dialogue Act Corpus) and MRDA (The ICSI Meeting Recorder Dialogue Act Corpus) are widely used Corpus in DA identification experiments. They have been widely accepted by scholars, and can be utilized as the benchmark dataset to evaluate the performance of DAC. Our UIIM model was also simulated on these two data sets. Taking the commonly used accuracy as the evaluation index, the value of SWDA and MRDA reached 78.6% and 89.9%, respectively, which is the state-of-the-art result at present.

The main contributions of this paper are as follows:

1. Based on the linguistic characteristics of utterance, this paper for the first time fully expounds the role of various kinds of cues in the task of dialogue act classification, and utilizes them as features to complete the recognition of DA.
2. In the field of DA classification, we proposed a new model, UIIM, to mine the universalities and individualities among various features, and achieved the best accuracy on two benchmark data sets, SwDA and MRDA.

## 2  Related Work

In recent years, both traditional machine learning and deep learning methods have achieved good performances in the classification of DAs. In this section, we will focus on introducing the methods that are similar to our research.

For DA classification task, the earliest representative research work is that Grau *et al.* utilized Naive Bayesian(NB) Classifier to realize acts recognition and evaluated it on the benchmark dataset SwDA, and the accuracy rate reached 66% [10]. Then, many scholars proposed to use structured algorithms to predict classification. For example, Stolcke *et al.* used Hidden Markov Model (HMM) to predict the speech's illocutionary force [11]. Tavafi *et al.* combined Support Vector Machine (SVM) with HMM, which further improved the accuracy of prediction [12, 13].

Recently, researchers have begun to introduce deep learning models to solve the problem of DA classification, and the accuracy has been greatly improved compared to traditional machine learning methods. Lee *et al.* realized that many short texts appear in order, so they proposed to use Convolutional Neural Network (CNN) and Recurrent Neural Network (RNN) to extract contextual information [4]. This greatly improves the classification performance of the model. Similarly, Kalchbrenner *et al.* also used CNN to extract sentence information, then used RNN to extract discourse information, and integrated full-text information to complete DA recognition [5]. Ortega *et al.* not only used RNN to extract discourse information, but also added the attention mechanism into the model, so that the model could extract the most relevant information more accurately and ignore some irrelevant information [14]. In addition to the above models which only consider the current discourse and context information simply, there are some models which preliminarily extract other feature information. Ortega *et al.* then fuse prosody information into the DA classifier and further improved the model performance [2].

## 3  Methodology

The goal of this paper is to identify the act of utterance (*U*) by using multiple clues. Every conversation in the data is divided into its constituent utterance, each utterance has its illocutionary force to express, and this utterance is also the smallest sample of recognition act. The low-level feature input of our model comes from three cues, including word, POS, and statistics. Let's denote them as expression (1).

$$U_w \in R^{l \times d_w}, U_p \in R^{l \times d_p}, U_s \in R^{d_s} \tag{1}$$

Here, $d_{f \in \{w, p, s\}}$ denotes the dimension of each feature *f* embedding, and *l* denotes the length of the utterance. For the specific information of these vectors, see Section 4.3. Through these utterance features $U_{f \in \{w, p, s\}}$, we can predict the act category $y \in R^{|C|}$ to which the utterance belongs, where *C* is the set of utterance act categories.

### 3.1  Model Architecture

In this paper, a new matrix containing current utterance information is obtained by feature extraction and integration. Then, the matrix gathers context information through BiLSTM. Finally, the act category of the utterance is obtained after processing by the fully connected network. Feature extraction and integration are the key of our model, which is introduced in two stages: one is representation learning, which extract all kinds of cues contained in utterance, mine their universality and individuality and represent them as vectors that can be recognized by computer. The other is feature fusion in order to integrate these vectors reasonably and input them into classifiers for prediction, shown in 错误!未找到引用源。.

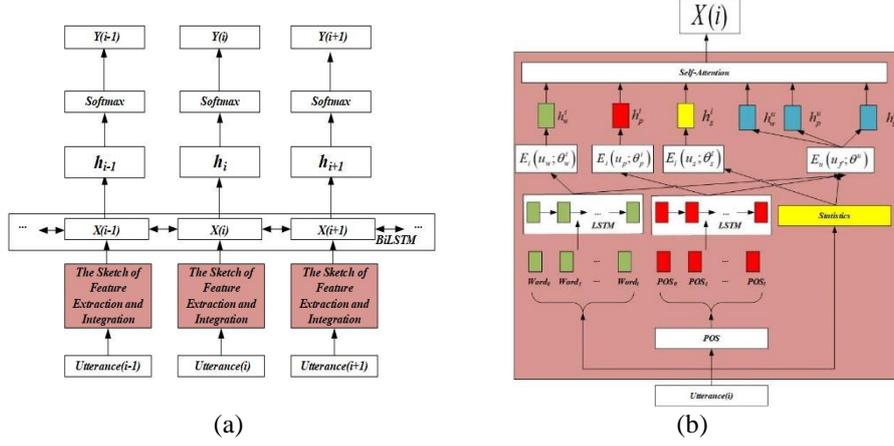

**Fig. 1.** Overview of UIIM architecture (Fig.1.a), where Feature Extraction and Representation is the key module of the model whose details are shown in Fig.1.b.

**Representation Learning.** In fact, the initial shapes of the vectors embedded by different features are varied[7]. In order to make the model compare the similarity between features more efficiently, we need to map the vectors to the same dimension[15]. Specifically, for each feature $f \in \{w, p, s\}$, if its initial vector is expressed as $U_f \in R^{l \times d_f}$ (when $f = s$, $U_s \in R^{d_s}$), we obtain new hidden vector $u_f^0 \in R^{d_f}$ after integrating the information of the word segmentation with LSTM, and then obtain the vector $u_f \in R^{d_h}$ of uniform dimension dh after transformation through the fully connected network. The above operation formula is expressed as equations (2) and (3).

$$u_f^0 = LSTM(U_f; \theta_f^{lstm}) \tag{2}$$

$$u_f = tanh(W_f^0 u_f^0 + b_f^0) \tag{3}$$

Where $\theta_f^{lstm}$ is the parameter of each $U_f$ in LSTM. $W_f^0$ is a weight matrix with a shape of $d_h \times d_f$, and b is a vector representing bias.

Next, we need to map each utterance vector $U_f$ into two different hidden vector representations, namely, universality vector $h_f^u$ and individuality vector $h_f^i$, as shown in Eq. (4) and (5). The former is a universal vector used to constrain the similarity of distribution, which is beneficial to deepen the relationship between features and highlight the commonality of vectors. The latter represents a feature-specific vector that is used to learn what is unique about that feature. In this paper, we believe that there are connections and gaps between features, but they are not superficial and transparent to us. Our task is to discover hidden universalities and individualities by studying these representations, which is shown in equations (4) and (5)

$$h_f^u = E_u(u_f; \theta^u) \tag{4}$$

$$h_f^i = E_i(u_f; \theta_f^i) \tag{5}$$

Here, the encoder function $E_u$ is used to generate the hidden vector $h_f^u$, and the function $E_i$ is used to generate the vector $h_f^i$. The encoder is implemented by feedforward neural network. $\theta^u$ is the shared parameter used in the universality coding of all features, and $\theta_f^i$ is the parameter allocated separately according to the features.

**Feature Fusion.** After the universality and individuality vectors of the three features are obtained by projection, a fusion mechanism is designed to better connect the six vectors[16]. In this paper, we choose to utilize the self-attention mechanism to achieve the fusion of features.

First, a matrix can be formed from the vectors, and then a self-attention operation will be performed on the matrix $M = [h_w^u, h_w^i, h_p^u, h_p^i, h_s^u, h_s^i]$, so that each vector can extract potentially helpful information from the other vectors to identify the dialog act. Feature fusion can be expressed as equations (6) to (9).

$$Attention(Q, K, V) = softmax\left(\frac{QK^T}{\sqrt{d_h}}\right)V \tag{6}$$

$$head_i = Attention(QW_i^Q, KW_i^K, VW_i^V) \tag{7}$$

$$MultiHead(Q, K, V) = Concat(head_1, \ldots, head_t)W^O \tag{8}$$

$$X = MultiHead(M, M, M) \tag{9}$$

Where given a matrix of query vectors $Q$, keys $K$ and values $V$, the scaled dot-product attention computes the attention scores based on the Eq. (6). The multi-head attention mechanism first maps the matrix of input vectors $I$ to matrices $Q$, $K$ and $V$ by using different linear maps. Formally, for the *i-th* head $head_i$, $W_i^Q$, $W_i^K$ and $W_i^V$ denote the learned maps, which correspond to queries, keys and values respectively. we set $Q = K = V = M$, then the attention generates a new matrix $X$. it is the combination of the universality and individuality of the three clues. Then a new vector is obtained by collecting contextual information through BiLSTM, which can be input into Multi-Layer Perceptron (MLP) to obtain the act category of this utterance[17].

## 3.2 Training

Then, we present the model UIIM implementation and training details. Here, the key operation is to minimize the loss function, which is shown in the equation (10).

$$Loss = \alpha Loss_{cls} + \beta Loss_u + \gamma Loss_i \quad (10)$$

Here, $Loss$ is the overall Loss value of the whole training process of the model. Theoretically, it contains three parts, that is, the classified Loss $Loss_{cls}$ of the distance between the predicted value and the real one. When training subtasks to obtain universalities of features, the loss $Loss_u$. And the third is the loss $Loss_i$ that happens when individualities is acquired. $Loss_u$ and $Loss_i$ were utilized for training to obtain the optimal universality vector $h_f^u$ and individuality vector $h_f^i$, respectively. Besides, $\alpha$, $\beta$ and $\gamma$ are the weights of the three loss functions relative to the overall, which can be determined experimentally.

**Universality Training.** In terms of the training of universalities, the scheme selected in this paper is as follows. Firstly, the initial universality vectors of the three features are encoded by Eq. (4). Then any two of them are taken as a group in turn to compare the similarity between them. Finally, the target vector with the highest similarity is determined by continuous training. Here, we define the loss function to achieve the optimal result by reducing the loss value. The loss function is shown in equations (11) and (12).

$$Loss_u = \frac{1}{3} \sum_{(f1,f2) \in \{(w,p),(w,s),(p,s)\}} \left(1 - cosine\left(h_{f1}^u, h_{f2}^u\right)\right) \quad (11)$$

$$cosine(x, y) = < x/\|x\|_2, y/\|y\|_2 > \quad (12)$$

Where $cosine(,)$ is used to calculate the cosine similarity between two vectors, $<x, y>$ is the dot product of $x$ and $y$ and $\|z\|_2$ is the l2-norm of $z$. In fact, $Loss_u$ is calculated based on cosine similarity, which contains a linear transformation to change its value range to [0,2]. The higher the similarity, the lower the loss value. We know that the distribution of different features is different, in other words, the assignment of each feature is different. The cosine similarity is not sensitive to the absolute value, but more to distinguish the difference from the direction, which corrects the problem that the measurement standard between features is not uniform. In view of the gap in the distribution of the compared features, the loss function is used in the paper.

**Individuality Training.** Similar to universality training, the process of individuality training generally consists of three stages. First, the initial individuality vectors are encoded by Eq. (5). Then any two as a group, compare the similarity between them. Finally, after continuous training, the target vectors with the least similarity are obtained. Note that, unlike universality training, any individuality vector not only has to be compared with the other two, but also with its own corresponding universality vector. Here, the loss function is still constructed and defined as equation (13).

$$Loss_i = \frac{1}{6}\left(\sum_{(f1,f2)\in\{(w,p),(w,s),(p,s)\}}\left(1+cosine(h^i_{f1},h^i_{f2})\right)+\sum_{f\in\{w,p,s\}}\left(1+cosine(h^u_f,h^i_f)\right)\right) \quad (13)$$

Here, $Loss_i$ denotes individuality loss, and its value range is [0,2].

**Classification Training.** Finally, the paper defines a loss function that measures the difference between the actual dialog act and the predicted act. Since the DAC is a multi-classification problem, cross entropy loss is applied in this paper. The loss function is shown in equation (14).

$$Loss_{cls} = -\sum_{k=1}^{N}(y_k \, log \, \hat{y}_k) \quad (14)$$

Where $y_k$ represents the true act label and $\hat{y}_k$ is the predicted value.

## 4       Experiments

In this section, we describe the experimental details of our approach. The relevant source code was implemented, and now is publicly available on Github[1].

### 4.1    Dataset

We evaluate the performance of our model on two benchmark datasets used in studies for the DAC task, the Switchboard Dialogue Act Corpus (SwDA)[2] and the ICSI Meeting Recorder Dialogue Act corpus (MRDA)[3]. The dataset description is shown in table 1.

**Table 1.** Statistics related to DA datasets.

| Dataset | \|C\| | \|V\| | Training | Validation | Testing |
|---|---|---|---|---|---|
| MRDA | 5 | 10K | 51(76K) | 11(15K) | 11(15K) |
| SwDA | 42 | 19K | 1003(173K) | 112(22K) | 19(4K) |

Table 1 summarizes the configuration of our datasets, |C| is the number of DA classes, |V| is the vocabulary size. Training, Validation and Testing indicate the number of conversations (number of utterances) in the respective splits. SwDA is a well-known telephone speech corpus, consisting of about 2,400 two-sided telephone conversation

---

[1]   https://github.com/Bonphy/UIIM
[2]   https://github.com/cgpotts/swda
[3]   https://github.com/NathanDuran/MRDA-Corpus

among 543 speakers with about 70 provided conversation topics (gardening, crime, music, etc.). In this dataset, utterances are annotated with 42 mutually exclusive DA labels (Statement-non-opinion, Acknowledge, Statement-opinion, etc.), based on the SWBD-DAMSL annotation scheme. The SwDA corpus contains an official split, consisting of 1003, 112, and 19 conversations for training, validation and testing, respectively [4]. MRDA contains 72 hours of naturally occurring multi-party meetings, and then hand annotated with DAs using the Meeting Recorder Dialogue Act Tagset. The dataset is marked with 5 DAs, namely, statements (s), questions (q), floorgrabber (f), backchannel (b), and disruption (d) [4].

### 4.2   Baseline

This paper mainly selects some of the state-of-the-art models as baseline, as follows.

- NB: proposed by Grau *et al.*, Naive Bayesian Classifier is utilized to realize acts recognition [10].
- SVM-HMM: proposed by Tavafi *et al.*, A combination of Support Vector Machine and Hidden Markov Model [13].
- CNN-RNN: proposed by Lee *et al.*, An utterance level CNN followed by a conversation CNN, with softmax classifiers. The utterance and conversation layers only consider the current utterance and at most 2 preceding ones [4].
- RCNN: proposed by Kalchbrenner *et al.*, Hierarchical CNN on word embeddings to model utterances followed by a RNN to capture context, with a softmax classifier[5].
- RNN-attention: proposed by Ortega *et al.*, the work not only used RNN to extract discourse information, but also added the attention mechanism into the model, so that the model could extract the most relevant information more accurately and ignore some irrelevant information [14].
- CNN-prosody: The work input prosody information into the classifier and further improved the model performance [2].
- fastText only: A simple Feedforward Neural Network with the fastText embedding.
- BERT only: A basic DAC model based on BERT pre-training.

### 4.3   Feature Extraction

**Word Features.** In general, the utterance text in each sample constitutes the source of word features. In order to extract this feature, the paper applies the pre-trained fastText embedding of 300-dimension to obtain the representation of words as vectors. Next, all the words in a sample are encoded by LSTM to obtain the final utterance representation. In addition, we also utilized the pre-train BERT to replace fastText as the feature extractor. Combined with relevant literature, we finally represent utterance with 768-dimension hidden state vector.

**POS Features.** On the dataset SwDA, POS has been tagged and can be obtained directly. As for MRDA, there is no ready-made POS label, so we can use the existing POS tagging tools, such as Natural Language ToolKit (NLTK) to quickly complete the tagging. Then, the part of speech of each word is expressed as one-hot, and all part of speech information is integrated with LSTM.

**Statistics Features.** We design a fixed-length row vector to collect statistical information, which can be roughly divided into two types: punctuation and utterance length. For punctuation, if the punctuation is included in the utterance, the corresponding element value is 1 in the vector; otherwise, it is 0. Besides, if the length is classified by interval, for example, the interval size is 5, then its set is $\{[0,4],[5,9],[10,14],[15,\infty)\}$. Then, we select the interval of the current utterance and mark it into the vector according to its length. Finally, we gain a vector whose dimension is equal to the sum of the number of punctuation categories and the size of the length set.

### 4.4 Hyperparameter Tuning

To find effective hyperparameters, we change one at a time while keeping the other ones fixed. Conversations with the same number of utterances were grouped together into mini-batches, and each utterance in a mini-batch was padded to the maximum length for that batch. The batch size chosen was 64. Dropout is applied to prevent overfitting of the model, and the final value was 0.3. The activation function ReLU was introduced to maximize the learning ability of the network The learning rate was initialized to 0.0001. Hidden size refers to the number of hidden layer neurons, which was set to 224. After the experiment, 1/0.7/0.7 was a reasonable choice for $\alpha/\beta/\gamma$.

### 4.5 Results

For both SwDA and MRDA, we apply the most popular split for training, validation, and testing as described in Section 4.1 above, so we are able to perfectly compare with the others. We present a comparison between different works and our model in Table 2.

Here, "fastText + UIIM" and "BERT + UIIM" are combined models based on embedding information and applying universality and individuality strategies. Comparing the two models with the baseline models, fastText only and BERT only, the combined models yield improvements over both base models on both datasets. It indicates that extracting part of speech features and statistical features to supplement word information, and then mining the dependencies and gaps between the three can improve the accuracy of recognition. Besides, the test performance of all the above models is the optimal result as mentioned in their literature. Our model outperforms the most advanced approach by 5.2% on MRDA, the primary data set for DA classification, and by 3.5% on SwDA. The improvement that this model can make over other methods is significant, which proves that it is more suitable for the recognition of dialogue act.

**Table 2.** Accuracy of our model and other models on SwDA and MRDA test sets.

| | Method | MRDA | SwDA |
|---|---|---|---|
| Baseline | fastText only | 88.5 | 75.6 |
| | BERT only | 89.2 | 77.9 |
| | NB | - | 66.0 |
| | SVM-HMM | 80.5 | 74.3 |
| | CNN-RNN | 84.6 | 73.9 |
| | RCNN | - | 73.9 |
| | RNN-attention | 84.3 | 73.8 |
| | CNN-prosody | 84.7 | 75.1 |
| Our Propose | fastText +UIIM | 89.3 | 76.4 |
| | BERT+UIIM | **89.9** | **78.6** |

## 5  Conclusions

In this paper, we proposed a classification model based on individuality and universality analyses to improve the information extraction and representation of word, part of speech and statistics. The model not only constructs the universality vectors of the clues by using the dependency relationship between them, but also pays attention to preserving the characteristics of the clues themselves and training their individuality vectors. Both of them comprehensively represent the feature information of utterance and provide a more rigorous basis for identifying acts. On the SwDA and MRDA data sets, the recognition accuracy of the model reaches 78.6% and 89.9%, respectively. The simulation results show that extracting the universality and individuality of features can more fully excavate the hidden information in utterances and improve the accuracy of automatic recognition of dialogue acts.